\pdfoutput=1

\documentclass[11pt]{article}

\usepackage[]{EMNLP2023}

\usepackage[normalem]{ulem}

\usepackage{times}
\usepackage{latexsym}

\usepackage[T1]{fontenc}
\usepackage[utf8]{inputenc}

\usepackage{microtype}

\usepackage{inconsolata}

\usepackage{algorithm}
\usepackage{algpseudocode}
\usepackage{tikz}
\usetikzlibrary{shapes.geometric, arrows, positioning}
\usepackage{booktabs}
\usepackage{multirow}
\usepackage{graphicx}
\usepackage{subcaption}

\usepackage{amssymb}
\usepackage{amsmath}
\usepackage{cleveref}
\usepackage{adjustbox}

\usepackage{enumitem}
\title{DecoMT: Decomposed Prompting for Machine Translation Between Related Languages using Large Language Models}

\author{Ratish Puduppully$^1$ \hspace{0.2cm}Anoop Kunchukuttan$^{3,5,6}$\hspace{0.3cm} Raj Dabre$^{4}$\hspace{0.3cm}\textbf{Ai Ti Aw}$^1$ \hspace{0.2cm} \textbf{Nancy F. Chen}$^{1,2}$\\\\
$^1$Institute for Infocomm Research (I$^2$R), A$^{*}$STAR, Singapore\\
$^2$CNRS@CREATE, Singapore\hspace{0.4cm}
    $^3$Microsoft, India\\
     $^4$National Institute of Information and Communications Technology\\
     $^5$IIT Madras\hspace{0.4cm} $^6$AI4Bharat\\     
\texttt{puduppully@i2r.a-star.edu.sg}~~~~\texttt{ankunchu@microsoft.com}~~~~\texttt{raj.dabre@nict.go.jp}\\ \texttt{nfychen@i2r.a-star.edu.sg}\\
}

\begin{document}
\maketitle
\begin{abstract}
This study investigates machine translation between related languages \textit{i.e.,} languages within the same family that share linguistic characteristics such as word order and lexical similarity. Machine translation through few-shot prompting leverages a small set of translation pair examples to generate translations for test sentences. This procedure requires the model to learn how to generate translations while simultaneously ensuring that token ordering is maintained to produce a fluent and accurate translation. We propose that for related languages, the task of machine translation can be simplified by leveraging the monotonic alignment characteristic of such languages. We introduce DecoMT, a novel approach of few-shot prompting that decomposes the translation process into a sequence of word chunk translations. Through automatic and human evaluation conducted on multiple related language pairs across various language families, we demonstrate that our proposed approach of decomposed prompting surpasses multiple established few-shot baseline approaches. 
For example, DecoMT outperforms the strong few-shot prompting BLOOM model with an average improvement of 8 chrF++ scores across the examined languages.
\end{abstract}

\section{Introduction}
In this work, we focus on the translation between related languages, a vital aspect from both economic and social perspectives. A considerable amount of commercial activity and social interaction occur between neighboring regions speaking two related languages. In these situations, pivot translation via a third language, such as English, can prove inefficient due to two inference steps which can also cause cascading errors \cite{DBLP:journals/csur/DabreCK20}. Instead, direct translation between related languages could significantly streamline trade and enhance social connections.

Related languages, often from the same family, share word order and lexical characteristics, leading to predominantly monotonic translations where word order is largely preserved. This is seen in languages like Hindi, Marathi, Malayalam, Tamil, Bengali, etc. from the Indian subcontinent, which follow a Subject-Object-Verb (SOV) structure. Similar monotonic translation relationships are also observed among other language pairs, such as Indonesian and Malay or Ukrainian and Russian.

Recent work has shown the power of few-shot prompting with large language models (LLMs) for tasks like machine translation, summarization, and question answering \cite{lin-etal-2022-shot,workshop2023bloom}. In machine translation, this approach prompts an LLM with a handful of example pairs and a test example. This requires the model to generate translations while ensuring a fluent word ordering, a process that fails to account for any unique characteristics intrinsic to the languages involved. For instance, it neglects the monotonic alignment—an integral trait evident in translations between related languages. 

LLMs are often biased towards English in their training data. For example, in mT5 \cite{xue-etal-2021-mt5}, Hindi and Malayalam tokens represent just 0.8\% and 0.07\% respectively. This imbalance hinders LLM performance in tasks involving non-English languages and English to non-English translations \cite{lin-etal-2022-shot}. In particular, for few-shot translation tasks between related languages, these models may not have encountered sufficient data in these languages. Overcoming these limitations can be achieved by incorporating inductive biases about related languages.

Recently, \citet{khot2022decomposed} introduced an approach known as decomposed prompting. This technique dissects a complex task into simpler, more manageable subtasks, each of which is addressed through few-shot prompting of LLMs.

\begin{figure}[t]
    \small
    \centering
    \includegraphics[width=0.48\textwidth]{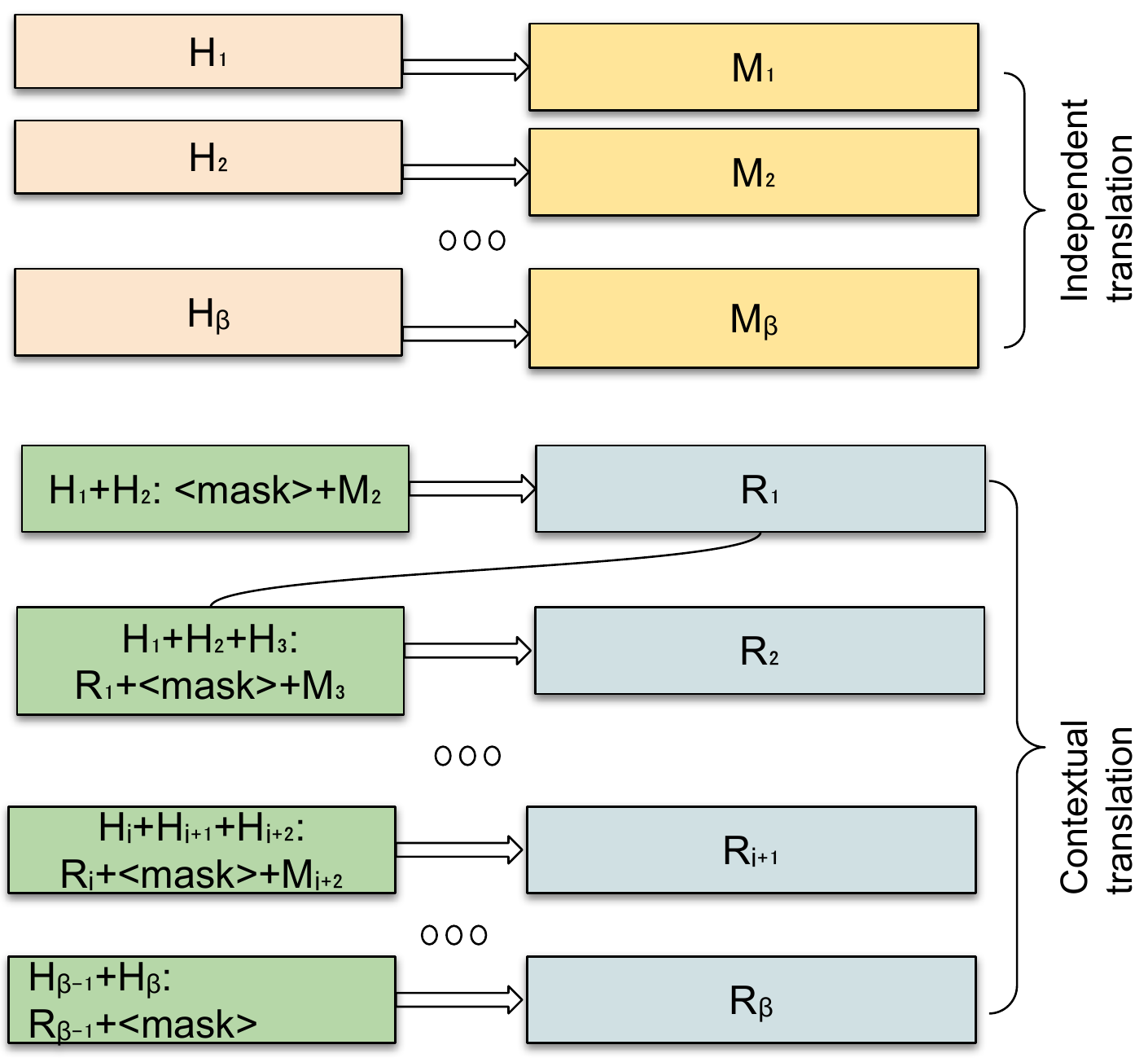}
    \caption{The diagram provides an overview of Decomposed Prompting for Machine Translation (DecoMT). The source text (H) is divided into several chunks (H$_1$, H$_2$,..,H$_i$,H$_{i+1}$,H$_{i+2}$,..,H$_{\beta}$). Each chunk is translated independently using few-shot prompting, yielding corresponding target chunks (M$_1$, M$_2$,..,M$_i$,M$_{i+1}$,M$_{i+2}$,..,M$_{\beta}$). The DecoMT process leverages the source chunks, their respective translations, and the previously predicted contextual translation to incrementally predict the contextually appropriate translation of the subsequent chunk.} %
    \label{fig:block_diagram_overview}
\end{figure}

We aim to enhance translations by harnessing the inductive bias of monotonicity in related languages. We posit that by relieving LLMs from implicit reordering and focusing on sub-sentence structures, more accurate translations, particularly in longer sentences, can be achieved. This leads us to propose a decomposed prompting approach, termed Decomposed Prompting for Machine Translation (DecoMT) (\Cref{fig:block_diagram_overview}), which splits an input sentence into chunks, translates each independently, and incrementally generates context-aware translations.

While much of the existing research on prompting focuses on decoder-only LLMs, recent studies \cite{patel2023bidirectional} show the potential of encoder-decoder models like mT5 \cite{xue-etal-2021-mt5} for such tasks. Our DecoMT approach builds upon this premise, utilizing the mT5 encoder-decoder LLM. 

The following are our contributions:

\begin{itemize}
    \item We introduce Decomposed Prompting for MT (DecoMT), a novel approach that simplifies the translation task by dividing it into the incremental translation of word chunks.
    \item We perform extensive evaluations on closely related languages from diverse language families, including pairs such as Hindi $\leftrightarrows$ Marathi, Hindi $\leftrightarrows$ Malayalam, Hindi $\leftrightarrows$ Telugu, Hindi $\leftrightarrows$ Gujarati, Indonesian $\leftrightarrows$ Malay, Russian $\leftrightarrows$ Ukrainian, and Spanish $\leftrightarrows$ Portuguese.
    \item We compare DecoMT against several robust baselines, including few-shot prompting of LLMs  \cite{lin-etal-2022-shot,workshop2023bloom}, as well as sequential autoregressive prompting of bidirectional LLMs \cite{patel2023bidirectional}. We demonstrate that DecoMT delivers robust results when compared to these baselines, particularly outperforming them in scenarios involving low-resource languages.
\end{itemize}

We release code and model outputs on github \footnote{\url{https://github.com/ratishsp/DecoMT}}.

\section{Related Work}

\paragraph{Few-shot Prompting for MT}
Few-shot prompting for MT leverages an autoregressive LLM, which is prompted with a small number of sentence pairs alongside their translations. The LLM then predicts the translation when provided with a test sentence. Examples of such LLMs include XGLM \cite{lin-etal-2022-shot} and BLOOM \cite{workshop2023bloom}. We interchangeably refer to this approach as Standard Prompting.

\citet{DBLP:conf/icml/GarciaBCFKJF23} have shown the effectiveness of few-shot prompting in machine translation. Yet, their method necessitates training a decoder-only LLM from scratch. In comparison, we use an off-the-shelf LLM, mT5, for DecoMT. A series of recent research delves into example selection for prompt construction \cite{vilar-etal-2023-prompting,pmlr-v202-zhang23m,kumar2023incontext,agrawal-etal-2023-context}. In our method, we rely on a fixed set of examples for prompting. \citet{jiao2023chatgpt} analyzed machine translation using ChatGPT and found that ChatGPT's performance aligns closely with commercial translation systems when utilizing GPT-4. In the interest of reproducibility, our emphasis lies on publicly accessible LLMs like BLOOM and mT5.

\paragraph{Sequential Autoregressive Prompting}
\citet{patel2023bidirectional} introduced an approach for prompting bidirectional LLMs, such as mT5 \cite{xue-etal-2021-mt5}. Their Sequential Autoregressive Prompting (SAP) method generates a token autoregressively, appends it back to the input, and predicts the subsequent token. They demonstrated that SAP outperforms traditional few-shot prompting for LLMs. Our method also leverages bidirectional LLMs. However, while they primarily exploit the autoregressive nature of these models, we further harness the bidirectional capability of LLMs to generate context-aware translations. %

\paragraph{Decomposed Prompting}
\citet{khot2022decomposed} proposed decomposed prompting, an approach that breaks down complex tasks into simpler ones, each tackled using few-shot prompting of LLMs. We apply this prompting strategy to the task of machine translation between related languages.

\paragraph{Incremental Generation}
In the field of data-to-text generation, \citet{puduppully-etal-2022-data} presented a strategy for document generation that decomposes the process into generating a sequence of paragraphs, interleaved with predicting a plan for each paragraph. Our DecoMT method can be viewed as an extension of this approach for the task of translating monotonically aligned sentences, where the plan is implicitly specified through the monotonic chunk alignment.

\citet{DBLP:journals/corr/abs-1810-13409} proposed an eager translation approach, in which the model begins translating without having to wait until the entire sentence has been processed. Our DecoMT method shares this characteristic, as it similarly doesn't require the whole sentence to be available before initiating translation. However, unlike their method, DecoMT's translation units extend beyond a single token. Moreover, DecoMT incorporates a contextual translation phase where the translation of an independent chunk is further refined through infilling.

\paragraph{Machine Translation for Low Resource Languages}
There have been studies on machine translation models for low-resource languages \cite{DBLP:journals/coling/HaddowBBHB22,nllbteam2022language,ramesh-etal-2022-samanantar,ai4bharat2023indictrans2,dabre-etal-2022-indicbart}. While most of these focus on translations between English and other languages, \citet{JMLR:v22:20-1307} is notable for its emphasis on improving translations among non-English languages. Our research aligns with this direction, concentrating on translations between related languages, many of which are characterized as low-resource.

\section{DecoMT}
In this section, we present the DecoMT Approach, our technique for decomposed prompting in Machine Translation.
Our method involves a two-stage translation process for word chunks: firstly, an independent translation stage where each chunk is translated in isolation; and secondly, a contextual translation stage where translation occurs while considering the surrounding context.

\subsection{Employed Pretrained Model}
In implementing DecoMT, we use the mT5 model \cite{xue-etal-2021-mt5}, specifically the XL variant with 3.7 billion parameters. mT5 is an encoder-decoder model that is trained with a span-corruption objective. During the training process of mT5, random spans within the input text are replaced with placeholders such as $\langle$mask\_0$\rangle$, $\langle$mask\_1$\rangle$, and so forth. In the output text, these correspond to mask tokens followed by the respective spans that were substituted in the input. 
Just like in the case of T5 \cite{DBLP:journals/jmlr/RaffelSRLNMZLL20}, the spans being replaced during training are of lengths varying from 2 to 5 tokens.

\begin{figure}[t]
    \centering
    \includegraphics[width=0.48\textwidth]{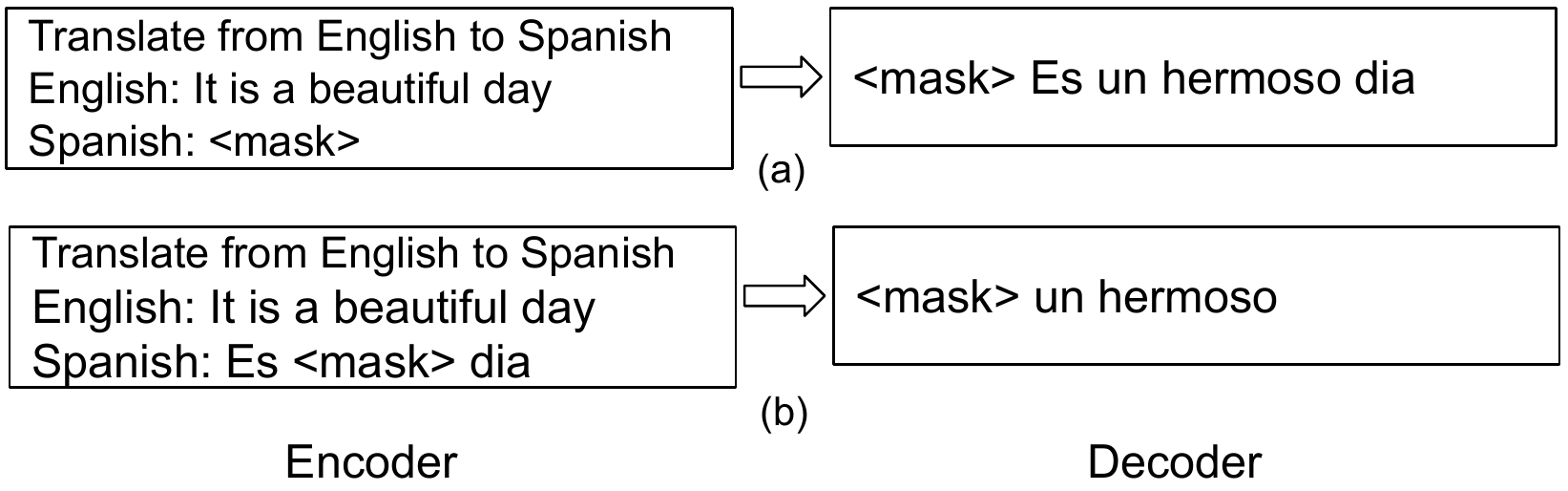}
    \caption{Depiction of two bidirectional encoder-decoder LLM prompting strategies for translation tasks. The upper part (a) uses an autoregressive translation, while part (b) employs the LLM for masked token infilling using surrounding context.}
    \label{fig:prompt=bidirection-lm}
\end{figure}

One approach to machine translation with mT5 follows the Standard Prompting method, as depicted in \Cref{fig:prompt=bidirection-lm} (a) \cite{workshop2023bloom,lin-etal-2022-shot}. %
In this setup, the mT5 encoder receives an input sequence: source language label, source sentence, target language label, followed by a $\langle$mask$\rangle$ token.
The decoder then generates the translation. In our independent translation framework, we employ this technique to produce M$_i$ from H$_i$, as depicted in \Cref{fig:block_diagram_overview}. 

Another technique to utilize mT5 for translation is by leveraging its bidirectional infilling capability, as exhibited in \Cref{fig:prompt=bidirection-lm} (b). %
The prompt includes the source language label, source sentence, target language label and a partially masked translation.
The mT5 decoder then generates the masked tokens. This specific approach is used in generating our contextual translations R$_i$ as shown in \Cref{fig:block_diagram_overview}. 

Depending on where the $\langle$mask$\rangle$ placeholder is inserted, the model will perform either text completion or infilling. It's important to note that a single mask can yield more than one token.

\subsection{Creating Aligned Monotonic Translations through Human Annotation}
We select the first five examples from the dev set of the FLORES dataset \cite{goyal-etal-2022-flores}. Each example consists of a pair of corresponding sentences in two different languages. Annotators are tasked to align these sentences in a monotonic manner, maintaining the same sequence of information. Importantly, annotators have the liberty to modify the sentences as required to achieve this.

\subsection{Translation Model}
Let $x$ represent the input sentence and $\beta$ denote the number of chunks in $x$. We define $\hat{y}$ as the preliminary translation of $x$, obtained by concatenating independently translated chunks. Furthermore, $y$ represents the final translation, which is assembled from contextually translated chunks.
For the purpose of simplification in our formulation, we omit the prompt template and focus on the translation of test examples.

In the case of independent translation, we make the assumption that each $\hat{y}_i$ is only dependent on its corresponding $x_i$, where $i$ indicates the index of the chunk within a sentence. This is captured by the equation:
\begin{equation}
p(\hat{y}|x) = \prod_{i=1}^{\beta} p(\hat{y}_i|x_i)
\end{equation}
In the case of contextual translation, we parameterise $y$ as dependent on $x$ and $\hat{y}$, represented as:
\begin{equation}
p(y|x,\hat{y}) = p(y_1 y_2 \dots y_\beta | x_1 x_2 \dots x_\beta, \hat{y}_1 \hat{y}_2 \dots \hat{y}_\beta)
\end{equation}
We make a conditional independence assumption that, at any position $i$, $y_i$ is dependent on $x_{i-1}$, $x_i$, $x_{i+1}$, the previous contextual translation $y_{i-1}$, and the next independent translation $\hat{y}_{i+1}$. This assumption allows us to rewrite the joint probability as a product of conditional probabilities:
\begin{align*}
p(y|x,\hat{y}) = & p(y_1 | x_1 x_2  \hat{y}_2) \\
& * \prod_{i=2}^{\beta - 1} p(y_i | x_{i-1} x_{i} x_{i+1} y_{i-1} \hat{y}_{i+1}) \\
& * p(y_{\beta} | x_{\beta -1} x_{\beta} y_{\beta - 1})
\end{align*}

\subsection{Prompt Construction}
Our methodology employs few-shot prompting, a technique that allows an LLM to make predictions based on a limited number of examples. This section will elucidate the process of constructing prompts for independent and contextual translation. We utilize five examples for few-shot prompting. 

\paragraph{Word count in Each Chunk}
Let us consider the token count within each word chunk in both prompt templates and test examples. For the prompt templates, $k$ and $j$ denote the number of tokens in a word chunk for independent and contextual translation, respectively. Conversely, in a test example, $m$ signifies the token count within a word chunk for independent translation.

We typically set $k$ and $j$ to 5 and 10, respectively. Nevertheless, the morphological richness of languages varies as a single token in one language might equate to several tokens in another. Hence, during the construction of prompt templates, we programmatically align each chunk fully with its translated equivalent, causing potential deviations from the standard values of 5 and 10 for $k$ and $j$.

Lastly, we treat $m$ as a hyperparameter, which is tuned using the FLORES development set.

\begin{figure}[t]
    \centering\includegraphics[width=0.48\textwidth]{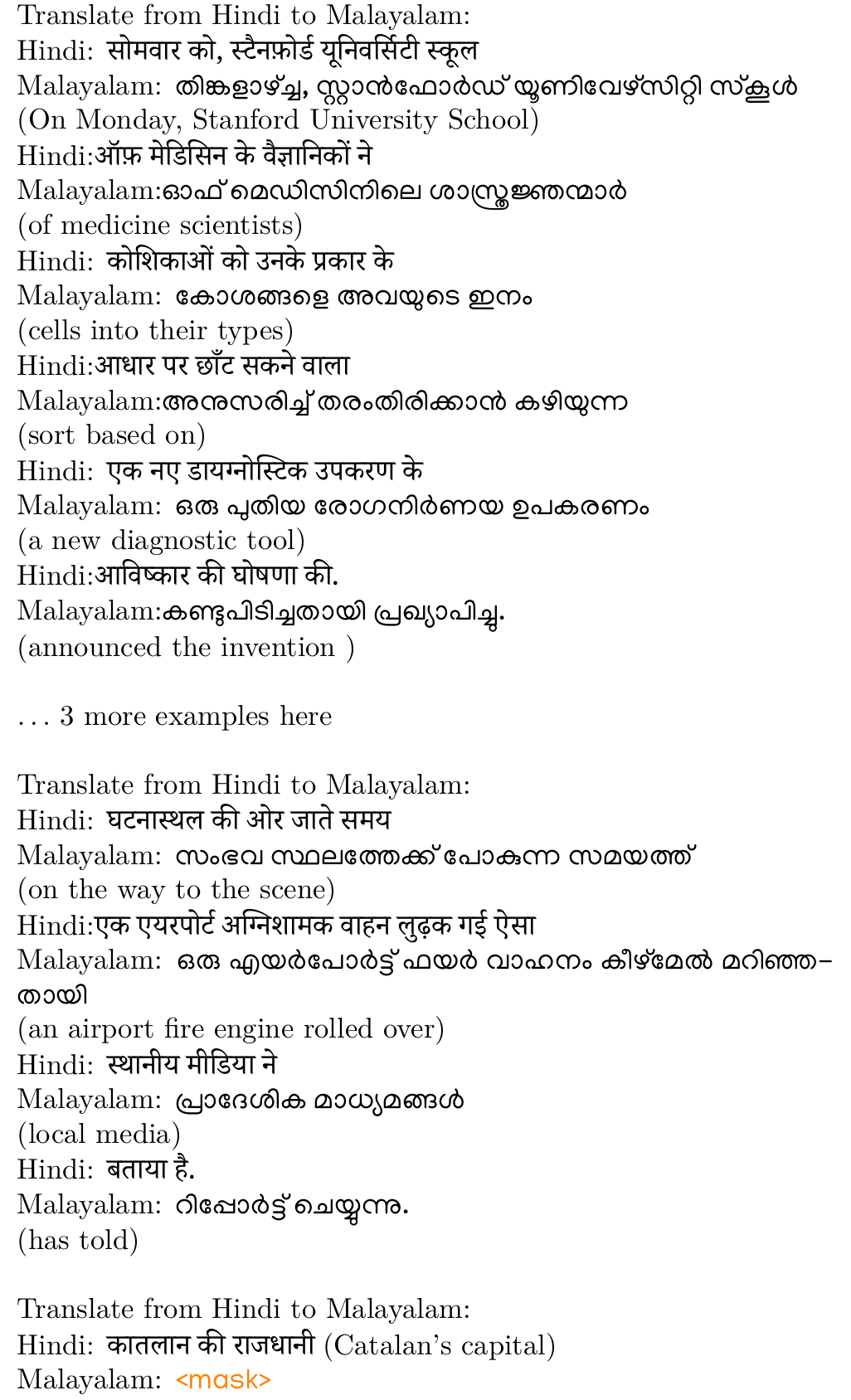}
\caption{Prompt Template for Independent Translation with a Test Example: The template includes five sentences in the source (Hindi) and target (Malayalam) languages divided into word chunks. The model receives a test example source chunk and a target language prompt with a $\langle$mask$\rangle$ placeholder, aiming to predict the corresponding target chunk. English text in brackets is for clarification, not in the actual prompt.}
    \label{tab:prompt_template}
\end{figure}

\begin{figure}[t]
    \centering
    \includegraphics[width=0.48\textwidth]{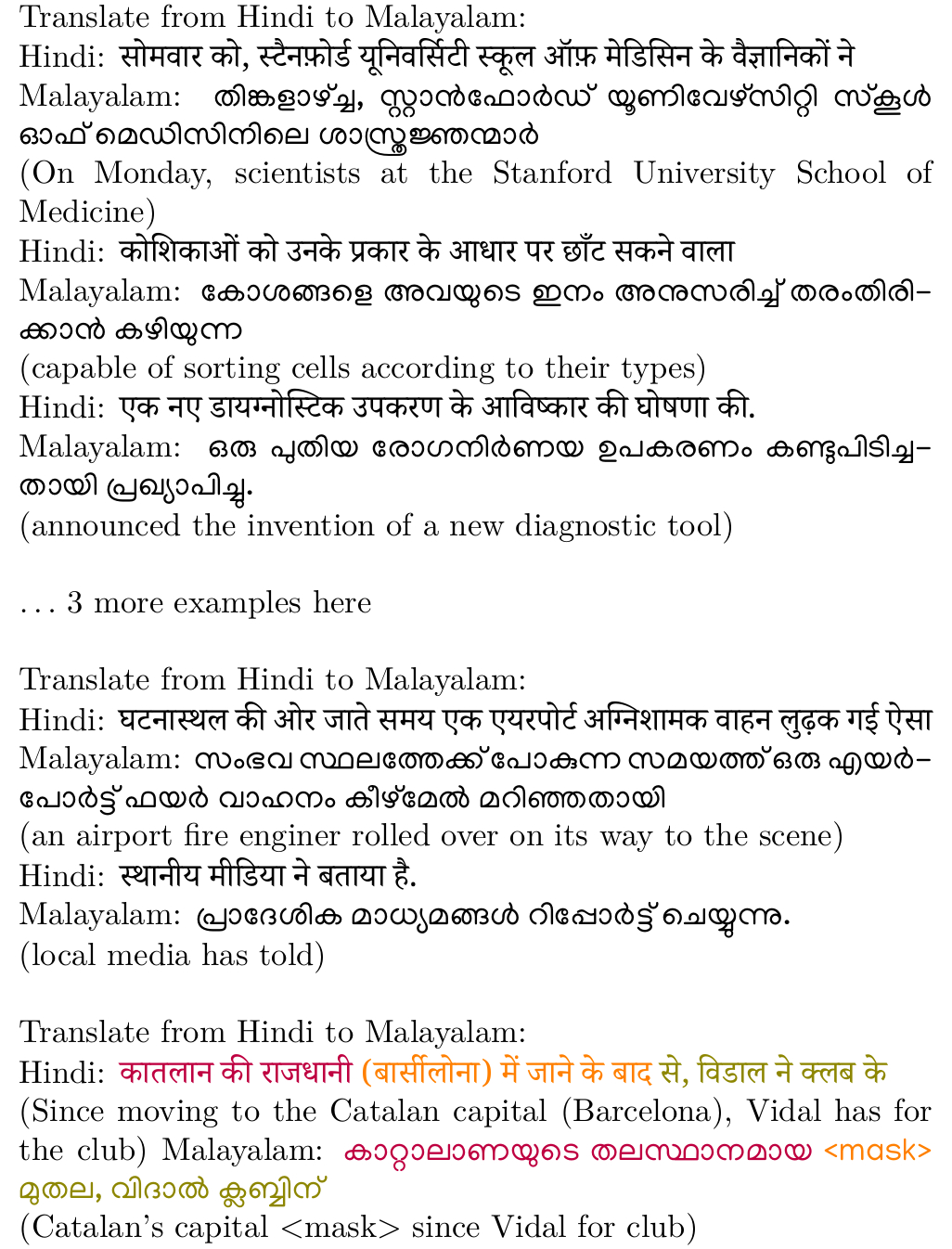}
    \caption{Prompt Template for Contextual Translation with a Test Example: Similar to \Cref{tab:prompt_template}, but with longer word chunks (approx. 10 tokens). The test prompt pairs a source language label with three concatenated word chunks. Following the target language label is the previous contextual translation, a $\langle$mask$\rangle$ placeholder, and the third chunk's independent translation. The model's goal is to complete the masked chunk. English bracketed text is explanatory and not a part of the prompt. The aligned chunks are colored identically.}
    \label{tab:revision_template}
\end{figure}
\paragraph{Independent Translation}
Each translation example for independent translation (\Cref{tab:prompt_template}) commences with ``Translate from [Source language] to [Target language]:'', followed by a line break, then ``[Source language]:'' and the first chunk of the source language sentence. Subsequently, we present ``[Target language]:'' and the corresponding translated chunk on a new line. This sequence is replicated for all the chunks in a sentence.

Upon completing a sentence, we use a newline separator and proceed to the next example. This procedure is repeated for all five examples in the prompt template.

In the case of the test example, the prompt begins with ``Translate from [Source language] to [Target language]:'', followed by a line break and ``[Source language]:'' with a chunk from the source language. The subsequent line is ``[Target language]: $\langle$mask$\rangle$''. The model's objective at this point is to predict the translation for the source language chunk.

\paragraph{Contextual Translation}
The prompt template for contextual translation (\Cref{tab:revision_template}) mirrors that of independent translation, with one key difference: the examples in prompt template are around twice as long as that of the lengths of examples in independent translation template prompt. 
In the test example for contextual translation, the prompt starts with ``Translate from [Source language] to [Target language]:'', followed by ``[Source language]:'' and a concatenation of three chunks from the source language.

The next line reads ``[Target language]: [previous contextual translation] $\langle$mask$\rangle$ [next independent translation]''. Here, the model's task is to infill the translation for the second source language chunk.

\Cref{app:prompt_template} contains an example of independent and contextual translation prompt templates for translation between Indonesian and Malay.
\subsection{Inference}
\Cref{fig:block_diagram_overview} provides an overview of our DecoMT approach. We omit the prompt template from the block diagram for simplicity. We segment the input sentence into multiple chunks, denoted as H$_1$, H$_2$, ..., H$_i$, H$_{i+1}$, H$_{i+2}$, ..., H$_{\beta}$, each comprising $m$ tokens. We then independently translate each chunk into corresponding translations, labelled as M$_1$, M$_2$, ..., M$_i$, M$_{i+1}$, M$_{i+2}$, ..., M$_{\beta}$.

The key innovation in our approach lies in the contextual translation, which is performed incrementally for each chunk. Initially, we concatenate the first two chunks, H$_1$ and H$_2$, with the placeholder $\langle$mask$\rangle$ and the translation of the second chunk M$_2$. This forms the input to predict the first contextual translation, R$_1$.

Subsequently, we concatenate the first three chunks, H$_1$, H$_2$, and H$_3$, with the contextual translation obtained from the previous step, R$_1$, alongside the placeholder $\langle$mask$\rangle$ and the translation of the third chunk, M$_3$. This is used to predict the next contextual translation, R$_2$.

This process is continued iteratively. At an intermediate step, the chunks H$_i$, H$_{i+1}$, and H$_{i+2}$, along with the previously computed contextual translation R$_{i}$, the placeholder $\langle$mask$\rangle$, and the translation of the chunk M$_{i+2}$, are used to predict the next contextual translation, R$_{i+1}$.

Finally, for the last chunk, the input is the concatenation of the penultimate and final chunks, H$_{\beta-1}$ and H$_{\beta}$, the last computed contextual translation R$_{\beta-1}$, and the placeholder $\langle$mask$\rangle$. The model then predicts the final contextual translation, R$_{\beta}$.

\Cref{sec:appendix_example} contains a worked out example for translation from Hindi to Malayalam. 

\begin{table*}[t]
    \centering
    \small
\begin{tabular}{lrrrrrrrrrr}
    \toprule
    {} & \multicolumn{5}{c}{spBLEU} & \multicolumn{5}{c}{chrF++} \\
    \cmidrule(lr){2-6} \cmidrule(lr){7-11}
    {} & \multicolumn{3}{c}{SP} & SAP & DecoMT & \multicolumn{3}{c}{SP} & SAP & DecoMT \\
    {} & {\tiny BLOOM} & {\tiny XGLM} & {\tiny mT5}& {\tiny mT5} & {\tiny mT5} & {\tiny BLOOM} & {\tiny XGLM} & {\tiny mT5} & {\tiny mT5} & {\tiny mT5} \\
    \midrule
    hin$\rightarrow$mal &  3.0 & 0.0  & 10.7 & \underline{17.6} & \bfseries 18.7 & 15.7 & 0.1 & 23.2  & \underline{34.3} & \bfseries 37.0 \\
    mal$\rightarrow$hin &  10.6 & 0.0 & 8.9 & \underline{14.9} & \bfseries 16.3 & 29.3 & 0.0 & 24.8 & \underline{34.2} & \bfseries 36.8 \\
    \midrule
    hin$\rightarrow$mar &  11.7 & 0.0 & 7.2& \underline{12.5} & \bfseries 13.9 & 30.8 & 2.8 & 22.4 & \underline{32.1} & \bfseries 35.6 \\
    mar$\rightarrow$hin &  \underline{19.7}$^{\tiny\dagger}$ & 0.0 &13.5 & 19.5 & \bfseries 21.0 & \underline{39.9} & 4.9 & 31.3 & 39.6 & \bfseries 41.9 \\
    \midrule
    hin$\rightarrow$guj &  6.8 & 0.0 & 15.3 & \underline{21.4} & \bfseries 22.0 & 26.2 & 0.1 & 30.9 & \underline{39.2} & \bfseries 41.1 \\
    guj$\rightarrow$hin &  20.8 & 0.0& 16.2  & \underline{22.5} & \bfseries 23.2 & 40.6 & 3.1 & 34.0 & \underline{42.2} & \bfseries 43.7 \\
    \midrule
    hin$\rightarrow$tel &  3.5 & 0.3& 9.2  & \underline{19.3}$^{\tiny\dagger}$ & \bfseries 19.5 & 19.9 & 1.6 & 24.0& \underline{37.2} & \bfseries 38.5 \\
    tel$\rightarrow$hin & 9.2 & 12.9& 9.6 & \underline{16.6} & \bfseries 17.8 & 28.7 & 30.6 & 26.2 & \underline{35.9} & \bfseries 38.6 \\
    \midrule    
    zsm$\rightarrow$ind &  -- & 0.0& 18.1  & \underline{28.7} & \bfseries 29.6 & -- & 7.4 & 40.8 & \underline{53.9} & \bfseries 55.9 \\
    ind$\rightarrow$zsm &  -- & 0.0 & 14.9 & \underline{26.9} & \bfseries 28.2 & -- & 3.4 & 37.2 & \underline{53.1} & \bfseries 54.5 \\
    \midrule
    rus$\rightarrow$ukr &  -- & 5.7 & 19.2 & \underline{30.1} & \bfseries 31.0 & -- & 24.3 & 36.4 & \underline{48.0} & \bfseries 49.9 \\
    ukr$\rightarrow$rus &  -- & 23.0& 17.8 & \underline{32.3} & \bfseries 34.4 & -- & 40.7 & 34.6 & \underline{49.1} & \bfseries 51.5 \\
\midrule
    spa$\rightarrow$por & \bfseries 29.1 & 28.3 & 13.6 & \underline{27.6} & 26.5 & \bfseries 51.5 & 49.4 & 32.0 & 48.4 & \underline{50.0} \\
    por$\rightarrow$spa & \bfseries 28.2 & 26.0 & 13.4 & 24.8 & \underline{26.3} & \bfseries 50.1 & 48.2& 33.1 & 46.4 & \underline{48.9} \\
    \bottomrule
\end{tabular}
\caption{The table presents spBLEU and chrF++ scores for standard prompting (SP) with BLOOM and XGLM, SAP with mT5, and our proposed DecoMT approach with mT5 across several language pairs, all tested on the FLORES devtest set. The highest performing results are highlighted in bold, and the second best scores are underlined for clarity. All comparisons with DecoMT demonstrate statistical significance ($p < 0.05$) (except results marked with~$^{\tiny\dagger}$) as per paired bootstrap sampling \cite{koehn-2004-statistical}.}
\label{tab:updated}
\end{table*}

\section{Experimental Setup}
We conduct a comparative study of our DecoMT approach, which is based on mT5 \cite{xue-etal-2021-mt5} with 3.7B parameters, against various established approaches. These include the Standard Prompting technique applied to 7.1B parameters variant of BLOOM \cite{workshop2023bloom}, and 7.5B parameters variant of XGLM \cite{lin-etal-2022-shot}.
We also compare our method with the Standard Prompting technique applied to the mT5 model. In this case, as mT5 generates only a few tokens at a time, we append the generated text back to the input to prompt further text generation.
Furthermore, we compare our approach with SAP \cite{patel2023bidirectional}, a technique that also utilizes mT5 with 3.7B parameters. %

\subsection{Evaluation Metrics}

Our approach's performance is assessed using spBLEU \cite{goyal-etal-2022-flores}, a variant of BLEU\cite{papineni-etal-2002-bleu}, and chrF++ \cite{popovic-2017-chrf} metrics. %
The BLEU metric measures word n-gram matches, encompassing unigram, bigram, trigram, and four-grams. However, due to the morphological richness of the languages we are working with, BLEU scores can often be underestimated. To counteract this, we employ spBLEU as suggested by NLLB \cite{goyal-etal-2022-flores,nllbteam2022language}, which utilizes a subword-based tokenizer.

Conversely, chrF++ evaluates character n-gram matches for n values ranging from 1 to 4, in addition to word n-gram matches that include unigram and bigram. Given its demonstrated higher correlation with human annotator scores for low-resource languages \cite{popovic-2017-chrf}, chrF++ serves as a valuable metric for our study.
We use the SacreBLEU library \cite{post-2018-call} to compute these metrics. %
We provide signatures for both BLEU \footnote{BLEU Signature: nrefs:1|case:mixed|eff:no|\newline tok:flores200|smooth:exp|version:2.3.1} and chrF++ \footnote{chrF++ Signature: nrefs:1|case:mixed|eff:yes|nc:6|\newline nw:2|space:no|version:2.3.1}. %

For hyperparameter tuning, we utilize the FLORES development set. We evaluate chunk sizes for $m$ from the set \{3,4,5\}. %

\subsection{Evaluation}
We conducted evaluations on multiple languages using the Flores devtest set, focusing specifically on translations between closely related languages: Hindi (hin) $\leftrightarrow$ Marathi (mar), hin $\leftrightarrow$ Malayalam (mal), hin $\leftrightarrow$ Gujarati (guj), hin $\leftrightarrow$ Telugu (tel), Indonesian (ind) $\leftrightarrow$ Malay (zsm), Ukrainian (ukr) $\leftrightarrow$ Russian (rus), and Portuguese (por) $\leftrightarrow$ Spanish (spa). The latter pair represents a high-resource language setup for comparison.

\section{Results}
\subsection{Automatic Evaluation}
The results of our evaluations are summarized in \Cref{tab:updated}. 
We conducted statistical significance testing via paired bootstrap sampling \cite{koehn-2004-statistical} ($p < 0.05$).
Regarding performance, XGLM \cite{lin-etal-2022-shot} when used with Standard Prompting, demonstrated low spBLEU and chrF++ scores for low-resource language pairs such as hin$\leftrightarrow$mal, hin$\leftrightarrow$mar, hin$\leftrightarrow$guj, and ind$\leftrightarrow$zsm. It performed somewhat better with the ukr$\rightarrow$rus pair, likely due to the greater availability of resources for Russian compared to Ukrainian.

BLOOM \cite{workshop2023bloom},  outperformed XGLM across all directions and language pairs except tel$\rightarrow$hin. However, BLOOM does not currently support languages such as zsm, rus, and ukr.

When implemented with Standard Prompting, mT5 outperformed XGLM for most low-resource language pairs and even outperformed BLOOM on hin$\rightarrow$mal, hin$\rightarrow$guj, and hin$\rightarrow$tel pairs, underscoring its effectiveness as a robust baseline.

SAP proved to be a strong approach, echoing the findings of \citet{patel2023bidirectional}. It outperformed Standard Prompting with BLOOM, XGLM and mT5 on the hin$\leftrightarrow$mal, hin$\leftrightarrow$mar, hin$\leftrightarrow$guj, hin$\leftrightarrow$tel, ind$\leftrightarrow$zsm, and rus$\leftrightarrow$ukr language pairs. Nevertheless, BLOOM outperformed SAP for the high-resource spa$\leftrightarrow$por pair.

Lastly, DecoMT surpassed all other approaches on the low-resource language pairs hin$\leftrightarrow$mal, hin$\leftrightarrow$mar, hin$\leftrightarrow$guj, hin$\leftrightarrow$tel, ind$\leftrightarrow$zsm, and rus$\leftrightarrow$ukr. While it also achieved impressive results with the high-resource spa$\leftrightarrow$por pair, it fell short of BLOOM's performance in this particular scenario.
It's worth noting that DecoMT demonstrated an average improvement of 13.8 points in the chrF++ score over Standard Prompting with mT5, which presents a more direct comparison for DecoMT due to the same base model and their similar prompting and inference strategies.

\subsection{Human Evaluation}
\label{sec:human_eval}
To further analyze the quality of the outputs and validate the enhancements indicated by the automatic evaluation scores, we carry out a human evaluation study. This involves a comparative examination of our DecoMT approach, SAP, and Standard Prompting with mT5 and BLOOM. 

We engaged annotators who possessed comprehension skills in the source language and demonstrated fluency in the target language. These annotators were remunerated in alignment with local hourly wage standards. The language pairs hin$\leftrightarrow$mar, hin$\leftrightarrow$guj, zsm$\rightarrow$ind, and por$\rightarrow$spa were selected for evaluation, contingent upon the availability of annotators well-suited for each pair. It should be noted that only a single annotator was assigned to each language pair. We sampled 50 sentences for each approach for a total of 200.

Our human evaluation strategy employs the Cross-Lingual Semantic Textual Similarity (XSTS) methodology \cite{licht-etal-2022-consistent} adopted by NLLB \cite{nllbteam2022language} and IndicTrans2 \cite{ai4bharat2023indictrans2}. Within this approach, annotators are presented with the source sentence alongside translations produced by various approaches, omitting any human-annotated references. As XSTS emphasizes translation adequacy over fluency, it is well-suited to our focus on translation between related, typically low-resource languages, where adequacy takes precedence.

The XSTS metric is composed of a scale ranging from 1 to 5, where a score of 1 signifies completely dissimilar sentence pairs and a score of 5 represents semantically identical sentences. \Cref{app:human_ann} contains details of the score values.

\begin{table}[t]
    \centering
    \small
    \begin{tabular}{lrrrr}
    \toprule
    {} & \multicolumn{2}{c}{SP} & SAP & DecoMT \\
    {} & {\tiny BLOOM} & {\tiny mT5}& {\tiny mT5} & {\tiny mT5} \\    
    \midrule
         hin$\rightarrow$mar& 2.4* & 1.9* &2.3* & 3.0 \\
         mar$\rightarrow$hin& 3.4 & 2.3* &3.4 & 3.6 \\
         hin$\rightarrow$guj &2.0* & 2.1*& 3.4&3.2 \\
         guj$\rightarrow$hin & 3.0*& 2.1*&3.3 &3.6 \\
         ind$\rightarrow$zsm&1.0*&3.4*&4.8&4.9\\          
         por$\rightarrow$spa&4.7&2.5*&4.1&4.5\\ 
         \bottomrule
    \end{tabular}
    \caption{Human evaluation scores for standard prompting (SP) with BLOOM and XGLM, SAP
with mT5, and our proposed DecoMT approach with mT5. Results marked with~* indicate a statistically significant difference ($p < 0.05$) from DecoMT using ANOVA with post-hoc Tukey HSD test.}
    \label{tab:human_eval}
\end{table}

As shown in \Cref{tab:human_eval}, DecoMT significantly outperforms Standard Prompting with mT5 across all language pairs. DecoMT is significantly better than BLOOM for hin$\rightarrow$mar, hin$\leftrightarrow$guj and ind$\rightarrow$zsm but comparable with BLOOM on mar$\rightarrow$hin and por$\rightarrow$spa. DecoMT is significantly better than SAP for hin$\rightarrow$mar, while demonstrating comparable performance for the remaining language pairs.

\section{Discussion}

\begin{figure*}[t]
    \centering
    \begin{subfigure}{0.45\textwidth}
        \centering
        \includegraphics[width=\linewidth]{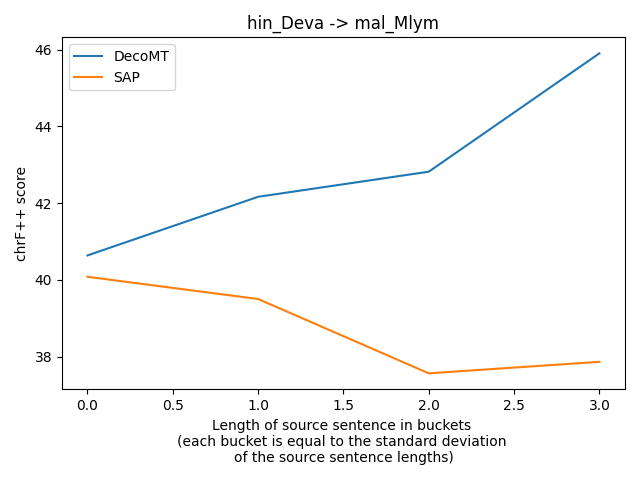}
    \end{subfigure}
    \begin{subfigure}{0.45\textwidth}
        \centering
        \includegraphics[width=\linewidth]{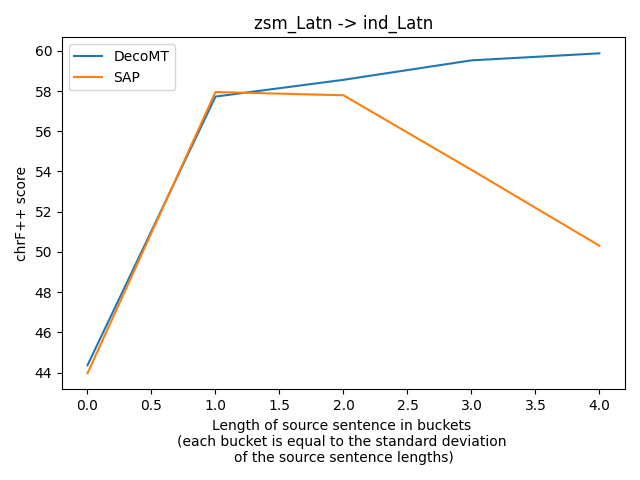}
    \end{subfigure}
    \caption{The plots show the relationship between source sentence length and chrF++ scores for hin$\rightarrow$mal and zsm$\rightarrow$ind pairs. Lengths are bucketed, each equal to the sentence lengths' standard deviation, with any bucket with less than 20 sentences merged with its neighbour. The data implies DecoMT's chrF++ scores outperform SAP's with increasing sentence length, indicating DecoMT's proficiency with longer sentences.}
    \label{fig:both_figures}
\end{figure*}

\paragraph{Scores of Translation across different Sentence Lengths}
The DecoMT strategy involves translating source sentences in consecutive chunks, a method we hypothesize will lead to enhanced translation adequacy. To explore this, we group source sentences into length-based buckets, each with a width equivalent to the standard deviation of the source sentence lengths. If a bucket contains fewer than 20 instances, we merge it with its neighbour. 

Figure \ref{fig:both_figures} depicts the relationship between source sentence length and chrF++ scores for the hin$\rightarrow$mal and zsm$\rightarrow$ind language pairs. As hypothesized, as the length of the source sentence increases, the performance of DecoMT, as measured by chrF++, improves.
For the zsm$\rightarrow$ind language pair, the chrF++ scores of DecoMT and SAP are nearly identical for the first two buckets. However, as we move to the next three buckets with longer sentences, we observe a steady increase in DecoMT's chrF++ scores. This is in contrast with the declining scores of SAP, highlighting DecoMT's superiority in translating longer sentences.

\paragraph{Improvement by Adding the Contextual Translation Compared to the Independent Translation}
We compared the single-stage independent translation to the two-stage DecoMT. The experiments show that the inclusion of contextual translation in the second stage of DecoMT significantly improves performance. We report the improvement in chrF++ scores in Table \ref{tab:translation_comparison_chrf}. The improvement in spBLEU is presented in \Cref{appendix:spbleu-improvement}.

\begin{table}[h]
    \centering
    \small
    \begin{tabular}{lcc}
        \toprule
        Lang Pair & chrF++ (Single Stage) & $\Delta$ chrF++ \\
        \midrule
        hin->mal & 33.7 & +3.3 \\        
        mal->hin & 34.7 & +2.1 \\\midrule
        hin->mar & 33.1 & +2.5 \\        
        mar->hin & 39.6 & +2.3 \\\midrule
        hin->guj & 39.6 & +1.5 \\
        guj->hin & 41.8 & +1.9 \\\midrule
        hin->tel & 36.3 & +2.2 \\
        tel->hin & 35.7 & +2.9 \\\midrule
        zsm->ind & 53.8 & +2.1 \\
        ind->zsm & 54.3 & +0.2 \\\midrule
        rus->ukr & 48.5 & +1.4 \\
        ukr->rus & 49.9 & +1.6 \\\midrule
        spa->por & 48.7 & +1.3 \\
        por->spa & 47.1 & +1.8 \\
        \bottomrule
    \end{tabular}
    \caption{Improvement in chrF++ scores gained by the DecoMT approach compared to the Single Stage.}
    \label{tab:translation_comparison_chrf}
\end{table}

\paragraph{Off-target Translations}
To quantify the off-target translation rate among various approach's outputs, we employed the Language Identification tool developed by the NLLB \cite{nllbteam2022language}. The off-target translation rate is represented as a percentage, with a lower percentage denoting superior performance, as shown in \Cref{tab:off_target}. We see that the DecoMT approach consistently outperforms other approaches with lower off-target translation rate across various translation tasks. We conduct further analysis in \Cref{app:off_target}.

\begin{table}[t]
    \centering
    \small
    \begin{tabular}{lrrrrr}
    \toprule
    {} & \multicolumn{3}{c}{SP} & SAP & DecoMT \\
    {} & {\tiny BLOOM}& {\tiny XGLM} & {\tiny mT5}& {\tiny mT5} & {\tiny mT5} \\    
    \midrule
         hin$\rightarrow$mal&23.6 &100.0&14.4&0.4&0.0 \\
         mal$\rightarrow$hin&8.4 & 0.0&4.4&1.4&0.2\\ \midrule    
         hin$\rightarrow$mar&21.2 &96.3&35.2&10.0&0.8 \\
         mar$\rightarrow$hin&1.3 &20.0 &2.6&1.1&0.2\\ \midrule
         hin$\rightarrow$guj &10.2&99.7 &3.8&0.2&0.0\\
         guj$\rightarrow$hin &3.3 &0.0 &1.9&0.4&0.2\\ \midrule
         zsm$\rightarrow$ind&--&48.8&23.3&17.7&13.1\\         ind$\rightarrow$zsm&--&94.2&59.7&47.3&30.1\\ \midrule
         rus$\rightarrow$ukr & -- & 84.3 & 1.7 & 0.2 & 0.0 \\ 
         ukr$\rightarrow$rus & -- & 0.6 & 0.5 & 0.1 & 0.0 \\\midrule
         spa$\rightarrow$por&0.2& 0.4&3.4&0.9&0.2\\ 
         por$\rightarrow$spa&0.0&0.5 &0.6&0.3&0.1\\ 
         \bottomrule
    \end{tabular}
    \caption{The percentage of sentences off-target for a translation direction. Lower is better.}
    \label{tab:off_target}
\end{table}

\paragraph{Extension to Autoregressive and other Encoder-Decoder LLMs}
At present, we utilize mT5 for both independent and contextual translations. However, it's worth noting that any autoregressive LLM could potentially be used for independent translation. As for contextual translation, an autoregressive LLM could be prompted with a fill-in-the-blanks type of prompt - an avenue we intend to explore in future work. Additionally, the exploration of other encoder-decoder LLMs such as UL2 \cite{tay2023ul2} or AlexaTM \cite{soltan2022alexatm} for contextual translations presents a promising research direction.

\paragraph{Experiments with Zero-shot and One-shot Prompting}
We undertook zero-shot translation experiments for select language pairs, specifically hin<->guj, hin<->tel, and hin<->mal. We compared different approaches applied to mT5 including DecoMT, SAP and Standard Prompting. We found that all approaches yielded near-zero BLEU scores. In most instances, the models merely copied the input as the output. We hypothesize that this is because in a zero-shot setting the model may not understand that it has to perform translation to the target language.

We compared one-shot and five-shot settings for three language pairs (hin<->guj, hin<->tel and hin<->mal) using Standard Prompting (SP), SAP, and DecoMT with mT5. Our results in \Cref{appendix:one-shot} indicate that:
\begin{itemize}
    \item DecoMT maintains strong performance even in the one-shot setting.
    \item Both SAP and SP experience significant performance drops transitioning from five-shot to one-shot. For instance, the spBLEU score for hin->tel in SAP drops from 19.3 (five-shot) to just 1.3 (one-shot). 
\end{itemize}

\paragraph{Inference Times}
As highlighted in \citet{patel2023bidirectional}, to generate a sentence comprising $T$ words, SAP necessitates $T$ forward passes through the model. This approach stands in contrast to Standard Prompting, which only requires a single pass. In the case of DecoMT, the independent translation stage can be parallelized with relative ease. For the contextual translation stage, $T/m$ forward passes through the model are needed, where $m$ denotes the chunk size. As a result, the inference time for DecoMT is less than that of SAP. \Cref{appendix:analysis_runtime} contains more details of runtime analysis.
\section{Conclusion}

In this study, we introduced DecoMT, a novel approach using decomposed prompting for Machine Translation of related languages. DecoMT demonstrated superior performance over established few-shot prompting baselines in translating between low-resource related languages, as evidenced by our experiments with the FLORES dataset. Additionally, DecoMT showed robust performance even in high-resource scenarios.

\section*{Limitations}

Despite its advantages, DecoMT does possess certain limitations. Notably, the approach requires human annotation for constructing the five example-aligned prompts in the template. However, our observations suggest that the annotators primarily need to modify existing translations, which is less laborious than generating translations from scratch, an activity that can be done in under 30 minutes. Conversely, other baseline approaches don't require such annotation and are able to directly utilize translation examples. 

When considering the translation time, DecoMT, given its two-stage process encompassing independent and contextual translations, inherently requires a longer duration to generate outputs compared to traditional few-shot prompting methodologies.

Another limitation of DecoMT is its dependency on an LM with infixing capabilities during the contextual translation stage. In the absence of infixing capabilities, this can be simulated on other LLM with appropriate prompting, and we plan to explore that in future work.

\section*{Ethics Statement}
This study does not involve any new data collection. We solely utilize publicly accessible datasets for conducting the experiments reported herein. Furthermore, for the purpose of annotation of translation examples and the human evaluation of machine translation outputs, we employ annotators who are duly compensated for their time and expertise, ensuring fair practices in line with established standards. 

\section*{Acknowledgements} We would like to thank the reviewers for their feedback. This research was supported by funding from the Institute for Infocomm Research (I2R) under A$^{*}$STAR ARES, Singapore. 
We extend our gratitude to Litton Kurisinkel, Aswanth Kumar, Siti Umairah, Ivan Kukanov, Swapnali Waghunde, and Fabian Ritter-Gutierrez for their work in annotating the few-shot prompts. Additionally, we'd like to thank Siti Umairah, Fabian Ritter-Gutierrez, Kunal Gandhi, and Faiz Masi for their contributions to the human evaluation experiments.
\bibliography{anthology,custom}
\bibliographystyle{acl_natbib}
\appendix

\section{Examples of Prompts}
\label{app:prompt_template}
The prompts used for independent and contextual translations by DecoMT for the language pair Malay$\rightarrow$Indonesian are presented in \Cref{tab:indpdt-zsm-ind} and \Cref{tab:context-zsm-ind}, respectively. Meanwhile, \Cref{tab:standard-zsm-ind} illustrates the prompts utilized for Standard Prompting and SAP.

\begin{table}[H]
    \centering
    \small
    \begin{tabular}{p{0.45\textwidth}}
    \toprule
Translate from Malay to Indonesian:\\
Malay: Saintis dari Stamford Universiti Sekolah\\
Indonesian: Ilmuwan dari Stanford University School of\\
Malay: Perubatan pada hari Isnin\\
Indonesian: Medicine pada hari Senin\\
Malay: mengumumkan penemuan alat diagnostik baharu\\
Indonesian: mengumumkan penemuan alat diagnostik baru\\
Malay: yang boleh menyusun sel berdasarkan\\
Indonesian: yang bisa mengurutkan sel berdasarkan\\
Malay: jenis: cip kecil dapat dicetak\\
Indonesian: tipe: cip kecil dapat dicetak\\
Malay: yang boleh dihasilkan menggunakan printer\\
Indonesian: yang bisa diproduksi menggunakan printer\\
Malay: inkjet standard dengan kos sekitar\\
Indonesian: inkjet standar dengan biaya sekitar\\
Malay: satu sen AS se cip.\\
Indonesian: satu sen AS per cip.\\
\\
Translate from Malay to Indonesian:\\
Malay: Ketua penyelidik mengatakan bahawa diagnosis\\
Indonesian: Ketua peneliti mengatakan bahwa diagnosis\\
Malay: ini mungkin dapat menghasilkan pengesanan\\
Indonesian: ini mungkin dapat menghasilkan deteksi\\
Malay: awal kanser, tuberkulosis, HIV, dan\\
Indonesian: dini kanker, tuberkulosis, HIV, dan\\
Malay: malaria kepada pesakit-pesakit di negara\\
Indonesian: malaria kepada pasien-pasien di negara\\
Malay: berpendapatan rendah, di mana kadar\\
Indonesian: berpenghasilan rendah, di mana tingkat\\
Malay: kesembuhan dari penyakit-penyakit seperti kanser\\
Indonesian: kesembuhan dari penyakit-penyakit seperti kanker\\
Malay: payudara boleh mencapai setengah dari\\
Indonesian: payudara bisa mencapai setengah dari\\
Malay: negara-negara kaya.\\
Indonesian: negara-negara kaya.\\
\\
Translate from Malay to Indonesian:\\
Malay: JAS 39C Gripen terhempas ke\\
Indonesian: JAS 39C Gripen jatuh ke\\
Malay: landasan sekitar jam 9:30\\
Indonesian: landasan pacu sekitar pukul 9:30\\
Malay: waktu tempatan (0230 UTC) dan\\
Indonesian: waktu setempat (0230 UTC) dan\\
Malay: meletup, mengakibatkan ditutup lapangan terbang\\
Indonesian: meledak, menyebabkan ditutupnya bandara\\
Malay: untuk penerbangan komersial.\\
Indonesian: untuk penerbangan komersial.\\
\\
Translate from Malay to Indonesian:\\
Malay: Juruterbang tersebut dikenalpasti sebagai Ketua\\
Indonesian: Pilot tersebut diidentifikasi sebagai Pemimpin\\
Malay: Pasukan Dilokrit Pattavee.\\
Indonesian: Skuadron Dilokrit Pattavee.\\
\\
Translate from Malay to Indonesian:\\
Malay: Media tempatan melaporkan sebuah kenderaan\\
Indonesian: Media lokal melaporkan sebuah kendaraan\\
Malay: pemadam api di lapangan terbang tergolek\\
Indonesian: pemadam api di bandara terguling\\
Malay: ketika dikendalikan.\\
Indonesian: saat sedang dioperasikan.\\
\\
Translate from Malay to Indonesian:\\
\bottomrule
\end{tabular}
    \caption{Prompt for Independent translation in DecoMT from Malay to Indonesian}
    \label{tab:indpdt-zsm-ind}
\end{table}

\begin{table}[H]
    \centering
    \small
    \begin{tabular}{p{0.45\textwidth}}
    \toprule
Translate from Malay to Indonesian:\\
Malay: Saintis dari Stamford Universiti Sekolah Perubatan pada hari Isnin\\
Indonesian: Ilmuwan dari Stanford University School of Medicine pada hari Senin\\
Malay: mengumumkan penemuan alat diagnostik baharu yang boleh menyusun sel berdasarkan\\
Indonesian: mengumumkan penemuan alat diagnostik baru yang bisa mengurutkan sel berdasarkan\\
Malay: jenis: cip kecil dapat dicetak yang boleh dihasilkan menggunakan printer\\
Indonesian: tipe: cip kecil dapat dicetak yang bisa diproduksi menggunakan printer\\
Malay: inkjet standard dengan kos sekitar satu sen AS se cip.\\
Indonesian: inkjet standar dengan biaya sekitar satu sen AS per cip.\\
\\
Translate from Malay to Indonesian:\\
Malay: Ketua penyelidik mengatakan bahawa diagnosis ini mungkin dapat menghasilkan pengesanan\\
Indonesian: Ketua peneliti mengatakan bahwa diagnosis ini mungkin dapat menghasilkan deteksi\\
Malay: awal kanser, tuberkulosis, HIV, dan malaria kepada pesakit-pesakit di negara\\
Indonesian: dini kanker, tuberkulosis, HIV, dan malaria kepada pasien-pasien di negara\\
Malay: berpendapatan rendah, di mana kadar kesembuhan dari penyakit-penyakit seperti kanser\\
Indonesian: berpenghasilan rendah, di mana tingkat kesembuhan dari penyakit-penyakit seperti kanker\\
Malay: payudara boleh mencapai setengah dari negara-negara kaya.\\
Indonesian: payudara bisa mencapai setengah dari negara-negara kaya.\\
\\
Translate from Malay to Indonesian:\\
Malay: JAS 39C Gripen terhempas ke landasan sekitar jam 9:30\\
Indonesian: JAS 39C Gripen jatuh ke landasan pacu sekitar pukul 9:30\\
Malay: waktu tempatan (0230 UTC) dan meletup, mengakibatkan ditutup lapangan terbang\\
Indonesian: waktu setempat (0230 UTC) dan meledak, menyebabkan ditutupnya bandara\\
Malay: untuk penerbangan komersial.\\
Indonesian: untuk penerbangan komersial.\\
\\
Translate from Malay to Indonesian:\\
Malay: Juruterbang tersebut dikenalpasti sebagai Ketua Pasukan Dilokrit Pattavee.\\
Indonesian: Pilot tersebut diidentifikasi sebagai Pemimpin Skuadron Dilokrit Pattavee.\\
\\
Translate from Malay to Indonesian:\\
Malay: Media tempatan melaporkan sebuah kenderaan pemadam api di lapangan terbang tergolek\\
Indonesian: Media lokal melaporkan sebuah kendaraan pemadam api di bandara terguling\\
Malay: ketika dikendalikan.\\
Indonesian: saat sedang dioperasikan.\\
\\
Translate from Malay to Indonesian:\\
\bottomrule
\end{tabular}
    \caption{Prompt for Contextual translation in DecoMT from Malay to Indonesian}
    \label{tab:context-zsm-ind}
\end{table}

\begin{table}[H]
    \centering
    \small
    \begin{tabular}{p{0.45\textwidth}}
    \toprule
Translate from Malay to Indonesian:\\
Malay: Pada hari Isnin, Saintis daripada Sekolah Perubatan Universiti Stamford mengumumkan penemuan alat diagnostik baru yang boleh mengasingkan sel-sel mengikut jenis: cip kecil yang boleh dicetak yang boleh dihasilakn menggunakan pencetak standard inkjet untuk kira-kira satu sen A.S setiap satu.\\
Indonesian: Ilmuwan dari Stanford University School of Medicine pada hari Senin mengumumkan penemuan alat diagnostik baru yang bisa mengurutkan sel berdasarkan tipe: cip kecil dapat dicetak yang bisa diproduksi menggunakan printer inkjet standar dengan biaya sekitar satu sen AS per cip.\\
\\
Translate from Malay to Indonesian:\\
Malay: Penyelidik utama mengatakan bahawa ia mungkin menghasilkan pengesanan awal kanser, tuberkulosis, HIV dan malaria kepada pesakit di negara-negara berpendapatan rendah, di mana kadar kemandirian untuk penyakit seperti kanser payu dara ialah separuh daripada di negara-negara yang lebih kaya.\\
Indonesian: Ketua peneliti mengatakan bahwa diagnosis ini mungkin dapat menghasilkan deteksi dini kanker, tuberkulosis, HIV, dan malaria kepada pasien-pasien di negara berpenghasilan rendah, di mana tingkat kesembuhan dari penyakit-penyakit seperti kanker payudara bisa mencapai setengah dari negara-negara kaya.\\
\\
Translate from Malay to Indonesian:\\
Malay: JAS 39C Gripen telah terhempas ke atas landasan sekitar jam 9:30 pagi waktu tempatan (0230 UTC) dan meletup, mengakibatkan lapangan terbang ditutup bagi penerbangan komersial.\\
Indonesian: JAS 39C Gripen jatuh ke landasan pacu sekitar pukul 9.30 waktu setempat (0230 UTC) dan meledak, menyebabkan ditutupnya bandara untuk penerbangan komersial.\\
\\
Translate from Malay to Indonesian:\\
Malay: Juruterbang telah dikenal pasti sebagai Ketua Pasukan Dilokrit Pattavee.\\
Indonesian: Pilot tersebut diidentifikasi sebagai Pemimpin Skuadron Dilokrit Pattavee.\\
\\
Translate from Malay to Indonesian:\\
Malay: Media tempatan melaporkan kenderaan api lapangan terbang terguling ketika memberi maklum balas.\\
Indonesian: Media lokal melaporkan sebuah kendaraan pemadam api di bandara terguling saat sedang dioperasikan.\\
\\
Translate from Malay to Indonesian:\\
\bottomrule
\end{tabular}
    \caption{Prompt for Standard Prompting and SAP from Malay to Indonesian}
    \label{tab:standard-zsm-ind}
\end{table}

\section{Example for DecoMT}
\label{sec:appendix_example}

\begin{figure*}[t!]
    \centering
    \includegraphics[width=\textwidth]{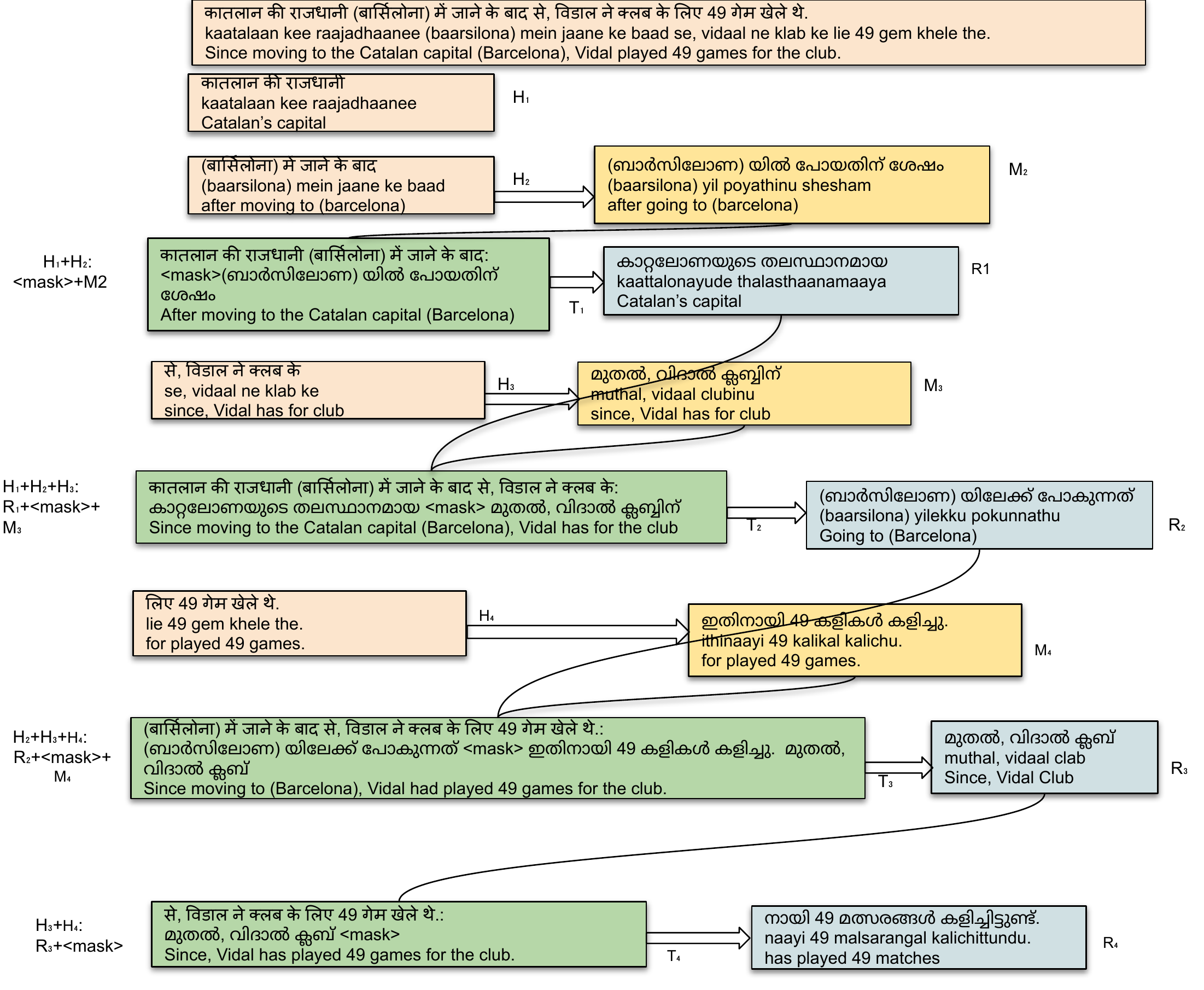}
    \caption{This diagram provides a step-by-step illustration of the DecoMT process. For the sake of simplifying our explanation, we have excluded the prompt template from the block diagram. The chunks of Hindi input, represented as H$_1$, H$_2$, H$_3$, and H$_4$, are initially translated into Malayalam independently using few-shot prompting, resulting in M$_1$, M$_2$, M$_3$, and M$_4$. Subsequently, infilling is used to derive contextual translations, denoted as R$_1$, R$_2$, R$_3$, and R$_4$. Each block of H$_i$, M$_i$, and R$_i$ presents three lines: the original text, its English transliteration, and its translation into English. The blocks marked T$_i$ illustrate the contextual translation tasks. The input block for T$_i$ includes a concatenation of input chunks, the previous contextual translation, a mask placeholder, and an independent translation, along with their English translation. The final translation into Malayalam, is produced by piecing together the contextual translations R$_1$, R$_2$, R$_3$, and R$_4$. It should be noted that the English translations and transliterations are included for the sake of clarity and are not an integral part of the DecoMT process.}
    \label{fig:block_diagram}
\end{figure*}

\Cref{fig:block_diagram} presents a block diagram which explains DecoMT with the help of an example. The task at hand is translation from Hindi to Malayalam. The Hindi sentence is divided into four consecutive chunks: H$_1$, H$_2$, H$_3$, and H$_4$, each consisting of $m=5$ tokens. Using few-shot prompting, these chunks are independently translated into Malayalam, resulting in M$_1$, M$_2$, M$_3$, and M$_4$. However, we observe that these translated chunks can occasionally lack coherence.

For instance, consider the translation of the H$_4$ chunk. The chunk commences with 
\adjustbox{valign=m}{\includegraphics[height=1.2em]{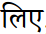}}
which can translate to `reason' or `for' (indicating possession) in English. The M$_4$ translation into Malayalam, 
\adjustbox{valign=b}{\includegraphics[height=1.2em]{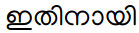}}
adopts the former meaning, whereas the sentence context implies that the latter interpretation would be more suitable.

To rectify this, we introduce a process to generate contextually appropriate translations. We input a concatenation of H$_1$, H$_2$, and a mask placeholder, along with M$_2$, into the bidirectional mT5 model. The model then infills the mask, producing a contextually appropriate translation of M$_1$, which we denote as R$_1$.

Next, we feed a concatenation of H$_1$, H$_2$, H$_3$, along with a concatenation of R$_1$, a mask placeholder, and M$_3$ into the mT5 model. The result is a contextually appropriate translation, R$_2$, of M$_2$.

This procedure is repeated for all the intermediate chunks. For the final chunk, we input a concatenation of H$_3$, H$_4$, R$_3$, and a mask placeholder. The mT5 model then predicts the contextually appropriate translation, R$_4$, of the M$_4$ translation. Given the context of H$_3$, H$_4$, and R$_3$, the contextual translation correctly interprets the intended meaning.

\section{Hyperparameter $m$}
The optimum value of $m$ for different language pairs is presented in \Cref{tab:m_hyperparam}. We posit that the optimal value of $m$ is contingent on the relative morphological complexity of the source language. Take the example of hin$\leftrightarrow$mal. Since Hindi (hin) is less morphologically complex than Malayalam (mal), a larger number of tokens are required in a chunk for hin$\rightarrow$mal than for mal$\rightarrow$hin to produce satisfactory outputs in the independent translation stage.

In the case of zsm$\leftrightarrow$ind, both languages exhibit similar morphological complexity, resulting in an identical optimum value of $m$, which is 4. The same applies to the rus$\leftrightarrow$ukr and spa$\leftrightarrow$por pairs. For these three pairs, a value of $m$ smaller than 4 results in subpar independent translation quality. Conversely, a value exceeding 4 might lead to truncated translations.

\begin{table}[t]
    \centering
    \small
    \begin{tabular}{cc}
    \toprule
         Language pair& $m$ \\
         \midrule
         hin$\rightarrow$mal&5 \\
         mal$\rightarrow$hin&3 \\\midrule
         hin$\rightarrow$mar&5 \\
         mar$\rightarrow$hin&4 \\\midrule
         hin$\rightarrow$guj&5 \\
         guj$\rightarrow$hin&4 \\\midrule
         hin$\rightarrow$tel&5 \\
         tel$\rightarrow$hin&3 \\\midrule     
         zsm$\rightarrow$ind&4 \\
         ind$\rightarrow$zsm&4 \\\midrule
         rus$\rightarrow$ukr&4 \\
         ukr$\rightarrow$rus&4 \\\midrule      
         por$\rightarrow$spa&4 \\
         spa$\rightarrow$por&4 \\         
         \bottomrule
    \end{tabular}
    \caption{Optimum value of m found through hyperparameter search in \{3,4,5\}.}
    \label{tab:m_hyperparam}
\end{table}

\section{Details of Human Annotation Guidelines}
\label{app:human_ann}
The XSTS metric provides ratings between 1 and 5, representing different levels of similarity between sentences.
\begin{itemize}
\item A score of 1 indicates that the sentences share little content or may be about different topics. If they share content, it is less than 50\%.
\item A score of 2 indicates that the sentences are about similar topics but are not equivalent, and there may be differences in important information related to the primary subject/verb/object.
\item A score of 3 indicates that the sentences are mostly similar, but there may be some minor omissions of unimportant information. There should not be any significant conflict in the information.
\item A score of 4 indicates that the sentences are paraphrases of each other. There are no major differences or missing information, although there may be variations in expression such as tone, style, emphasis, or formality.
\item A score of 5 indicates that the sentences are completely equivalent in meaning and usage, including expression aspects such as formality, tones, style, and emphasis.
\end{itemize}
For more details and examples,  see \citet{licht-etal-2022-consistent}.

\section{Improvement by Adding the Contextual Translation Compared to the Independent Translation}
\label{appendix:spbleu-improvement}

Table~\ref{tab:translation_comparison_spbleu} showcases the improvements in spBLEU scores achieved by the DecoMT approach in comparison to the Single Stage method.

\begin{table}[h]
\small
    \centering
    \begin{tabular}{lcc}
        \toprule
        Lang Pair & spBLEU (Single Stage) & $\Delta$ spBLEU \\
        \midrule
        hin->mal & 15.9 & +2.8 \\
        mal->hin & 13.3 & +3.0 \\\midrule
        hin->mar & 12.1 & +1.8 \\
        mar->hin & 17.0 & +4.0 \\\midrule
        hin->guj & 20.2 & +1.8 \\
        guj->hin & 21.0 & +2.2 \\\midrule
        hin->tel & 16.7 & +2.8 \\
        tel->hin & 11.3 & +6.5 \\\midrule
        zsm->ind & 26.4 & +3.2 \\
        ind->zsm & 27.7 & +0.5 \\\midrule
        rus->ukr & 28.3 & +2.7 \\
        ukr->rus & 32.1 & +2.3 \\\midrule
        spa->por & 24.4 & +2.1 \\
        por->spa & 23.7 & +2.6 \\
        \bottomrule
    \end{tabular}
    \caption{Improvement in spBLEU scores gained by the DecoMT approach compared to the Single Stage.}
    \label{tab:translation_comparison_spbleu}
\end{table}

\begin{table*}[ht]
    \centering
    \small
    \begin{tabular}{lcccccc}
        \toprule
        {} & \multicolumn{3}{c}{spBLEU} & \multicolumn{3}{c}{chrF++} \\
        \cmidrule(lr){2-4} \cmidrule(lr){5-7}
        Language Pair & SP & SAP & DecoMT & SP & SAP & DecoMT \\
        {} & {\tiny mT5} & {\tiny mT5} & {\tiny mT5} & {\tiny mT5} & {\tiny mT5} & {\tiny mT5} \\
        \midrule
        hin->guj one-shot & 10.0 & 16.1 & 22.2 & 20.5 & 29.6 & 41.1 \\
        hin->guj five-shots & 15.3 & 21.4 & 22.0 & 30.9 & 39.2 & 41.1 \\\midrule
        guj->hin one-shot & 17.1 & 22.9 & 23.0 & 34.7 & 42.7 & 43.4 \\
        guj->hin five-shots & 16.2 & 22.5 & 23.2 & 34.0 & 42.2 & 43.7 \\\midrule
        hin->tel one-shot & 0.4 & 1.3 & 18.9 & 1.5 & 2.8 & 38.2 \\
        hin->tel five-shots & 9.2 & 19.3 & 19.5 & 24.0 & 37.2 & 38.5 \\\midrule
        tel->hin one-shot & 5.3 & 8.7 & 17.4 & 12.6 & 18.6 & 38.4 \\
        tel->hin five-shots & 9.6 & 16.6 & 17.8 & 26.2 & 35.9 & 38.6 \\\midrule
        hin->mal one-shot & 1.3 & 2.9 & 18.2 & 3.2 & 5.7 & 36.7 \\
        hin->mal five-shots & 10.7 & 17.6 & 18.7 & 23.2 & 34.3 & 37.0 \\\midrule
        mal->hin one-shot & 9.1 & 13.4 & 16.5 & 22.7 & 29.8 & 36.9 \\
        mal->hin five-shots & 8.9 & 14.9 & 16.3 & 24.8 & 34.2 & 36.8 \\
        \bottomrule
    \end{tabular}
    \caption{Comparison of one-shot and five-shot translation results across three language pairs using SP, SAP, and DecoMT with mT5. Notably, DecoMT exhibits robust performance in one-shot settings, whereas SP and SAP show marked performance reductions, exemplified by the spBLEU drop for hin->tel in SAP from 19.3 (five-shot) to 1.3 (one-shot).}
    \label{tab:translation_results}
\end{table*}

\section{Off-target Translations}
\label{app:off_target}
In \Cref{tab:off_target}, focusing on the relatively high off-target translation rate for ind$\leftrightarrow$zsm, particularly for ind$\rightarrow$zsm, we analyzed 50 mislabeled DecoMT output sentences from ind$\rightarrow$zsm. An annotator from our human evaluation study (\Cref{sec:human_eval}) found that 64\% of these sentences were in fact Malay, not Indonesian. This suggests potential shortcomings in automatic language identification for closely related languages such as ind and zsm.

\section{Comparison between One-shot and Five-shot Prompting}
\label{appendix:one-shot}

As detailed in Table~\ref{tab:translation_results}, our evaluations span three language pairs and compare the efficacy of Standard Prompting (SP), SAP, and DecoMT methodologies when evaluated on mT5. In comparison between one-shot and five-shot scenarios, we find that DecoMT consistently demonstrates strong performance in one-shot settings, in contrast to the pronounced performance dips observed for both SP and SAP.

\section{Analysis of Runtime}
\label{appendix:analysis_runtime}
To ensure a fair comparison, we profile the codes using cprofile \footnote{\url{https://docs.python.org/3/library/profile.html}} during the inference phase, executed on an A40 48GB GPU. cprofile examines the time taken by various API calls. In this case, our chosen task is translating from Marathi to Hindi using the initial batch of 5 examples from the FLORES test set, with the longest Marathi sample in the batch being 41 tokens long.
\begin{itemize}
    \item SAP Analysis: For the SAP system, due to the unpredictability of the expected target length, we do decoding at 1.5 times the maximum source length. This is based on our studies of lengths of examples from validation dataset. For example, for our given source batch, the reference Hindi translation encompasses 55 tokens for the Marathi sentence which is 41 tokens long. As the longest example is 41 tokens, we run inference for 41 * 1.5 = 61 steps. \Cref{tab:performance-sap} contains a partial trace of performance profiling using cprofile.  We see that for SAP, there are 61 calls to predict\_output method. The method predict\_output is responsible for running inference on the LLM. Each method takes 2.384 seconds. The inference of the batch takes 145.455 seconds. 
    
    \begin{table}[h]
        \centering
        \small
        \begin{tabular}{|c|c|c|p{3.3cm}|}
            \hline
            ncalls & cumtime & percall & filename:lineno(function) \\
            \hline
            3/1 & 145.455 & 145.455 & \texttt{\{built-in method builtins.exec\}} \\
            \hline
            \ldots & \ldots & \ldots & \ldots \\
            \hline
            61 & 145.428 & 2.384 & \texttt{sap.py:163 (predict\_output)} \\
            \hline
        \end{tabular}
        \caption{Performance Profiling Data for SAP}
        \label{tab:performance-sap}
    \end{table}

    \item DecoMT Analysis: For Marathi-Hindi translations, we use a chunk size of 4. We first consider the independent translation stage. Breaking down the sentence lengths of the batch in tokens: 16, 30, 24, 41, and 28, we get respective chunk counts of 4, 8, 6, 11, and 7—aggregating to 36 chunks. Split into batches of 8, this leads to 5 API calls to predict\_output. With the longest sentence in the batch having 41 tokens, the contextual translation stage demands 11 API calls to predict\_output, cumulating to 16 calls. These 16 api calls in total amount to 96.868 seconds (\Cref{tab:decomt-profile}). While predict\_output in DecoMT tends to take longer than in SAP (owing to DecoMT predicting multiple tokens as opposed to SAP's single-token approach), the overall fewer API calls render DecoMT more efficient.

\begin{table}[h]
    \centering
    \small
    \begin{tabular}{|c|c|c|p{3.3cm}|}
        \hline
        ncalls & cumtime & percall & filename:lineno(function) \\
        \hline
        3/1 & 96.883 & 96.883 & \texttt{\{built-in method builtins.exec\}} \\
        \hline
        \ldots & \ldots & \ldots & \ldots \\
        \hline
        16 & 96.868 & 6.054 & \texttt{decomt.py:199 (predict\_output)} \\
        \hline
    \end{tabular}
    \caption{Performance Profiling Data for DecoMT}
    \label{tab:decomt-profile}
\end{table}
\end{itemize}

\end{document}